# State-of-the-art in Learning-by-Demonstration with Passive Observation for Industrial Assembly Automation

David Kötter*,[a], Oliver Petrovic[a], Christian Brecher [a,b]

[a]RWTH Aachen University, Laboratory for Machine Tools and Production Engineering, Steinbachstraße 19, 52074 Aachen, Germany
[b]Fraunhofer Intitute for Production Technology IPT, Steinbachstraße 17, 52074 Aachen, Germany

* Corresponding author. Tel.: +49-241-80-20224. *E-mail address:* d.koetter@wzl.rwth-aachen.de

**Abstract**

Learning-by-Demonstration (LbD) enables intuitive robot programming by capturing expert skills, which is crucial for agility in high-mix, low-volume manufacturing. This systematic literature review analyzes passive LbD for industrial assembly processes, focusing on the perception architecture and the generalization of the perceived demonstration. We specifically investigate one-shot approaches where only a single demonstration is required. The review evaluates how systems adapt to new assemblies using this limited data. We identify a shift towards object-centric perception, allowing learned primitives to be transferred to new product variants with minimal training.

## 1. Introduction and Background

The global manufacturing industry is shifting toward High-Mix, Low Volume (HMLV) production to address the rising demand for customized products [1, 2]. This transition is exacerbated by a systematic labor shortage in industrialized countries that threatens operational continuity [3] and by increasing supply chain volatility fueled by geopolitical tensions [4] and economic policy uncertainty [5]. To survive these pressures, manufacturers are moving away from rigid, dedicated automation lines in favor of flexible automation systems – centering on industrial and collaborative robots – that can be faster reconfigured to adapt to these circumstances [6]. However, traditional robot (re-)programming remains a significant bottleneck, often being too cost-intensive and technically complex for the frequent setup changes required in HMLV environments [7]. Consequently, Learning-by-Demonstration (LbD) has emerged as solution, providing a framework where robots acquire complex assembly skills by imitating human movements, thereby drastically reducing deployment costs and enabling agile automation in unpredictable market conditions [8]. LbD is structured into three main phases: a demonstration phase, a training phase, and an imitation phase [22]. The process begins with the demonstration phase, where a human operator performs the desired task, and the relevant data, including inputs and expected outputs, is captured. Following this, in the training phase, the data is processed and transformed into a plan (also called policy) how to solve the desired task. It often involves a machine learning algorithm processing the collected data to establish a mapping between inputs and outputs. Finally, in the imitation phase, the robotic system executes the learned behavior, applying the skills acquired during the previous stages. [9]

The implementation of LbD in industrial assembly is categorized by both the modality of data capture during the demonstration phase and the abstraction level of the learned information. To capture data, there are three different possibilities: kinesthetic teaching, teleoperation, and passive observation [10]. Kinesthetic teaching involves a human physically guiding the robot through desired movements, especially prevalent with collaborative robots (cobots). Teleoperation enables a human to remotely control a robot, proving valuable in environments hazardous or inaccessible to humans, or for tasks demanding fine manipulation. As both methods require the demonstrator to move the robot during the demonstration phase, production must be stopped while data is captured [11]. Lastly, passive observation involves robots learning manipulation skills by observing human actions, typically through video demonstration [10]. Using passive observation, there is no need to stop production in the demonstration phase, as external sensors (instead of the internal sensors of the robot) are used to capture data.

Regardless of the demonstration form, the system must determine *what* to learn, ranging from low-level trajectories (the specific path of motion) to symbolic encoding (the underlying task constraints and goals) [12]. While trajectory learning is suitable for repetitive tasks in static environments which do not require dynamic adaptation of the task execution, symbolic extraction enables a high degree of generalization. Moreover, a distinction can be made between the number of demonstrations required to capture data. Generally, a single demonstration is superior to multiple demonstrations, especially in a highly dynamic environment like the HMLV assembly. Therefore, the goal of this research is to evaluate the current literature in LbD by means of passive observation, learning high-level assembly tasks with a single demonstration. Section 2 introduces previous attempts to systematically capture the state of the art in LbD for assembly processes and introduces the research questions for this review. The research methodology is outlined in Section 3 and Section 4 gives the results of the literature review. Section 5 concludes this research with a summary and future lines of investigation.

## 2. Insights from previous literature reviews

To systematically evaluate and categorize existing literature, the scope fitness between previous reviews and the goal of this survey is determined in three dimensions: assembly (AS), passive observation (PO), and single demonstration (SD). In addition, the review method, the content of the research and the outlook are detailed in Table 1.

Tsuj et al. present a comprehensive survey of LbD methodologies specifically designed for contact-rich robotic tasks that address the inherent challenges of modeling complex physical interactions and nonlinear dynamics. They cover a variety of demonstration collection methodologies and learning approaches, emphasizing the need to fuse multimodal sensory action. [13] In contrast, Zare et al. systematically categorize LbD methodologies into two foundational approaches: behavioral cloning, which employs supervised learning to map states to expert actions, and inverse reinforcement learning, which focuses on recovering the latent reward functions optimized by the demonstrator. However, for all reviewed approaches, multiple demonstrations are required to obtain applicable results. [14] Urain et al. provide a comprehensive analysis of recent advances in deep generative models (DGM) for LbD, focusing on the transition from classical imitation learning methods to frameworks capable of capturing complex, multimodal data distributions. The research categorizes various DGMs – including diffusion models, energy-based models, and generative adversarial networks – and evaluates their applicability to robotic-specific challenges such as demonstration diversity, partial observability, and heterogenous action spaces. To train DGMs, multiple demonstrations are required. [15] Li et al. explore the current state and future directions of autonomous robotic trajectory generation through LbD by applying a two-step systematic literature review. They systematically categorize and benchmark primary demonstration modes, including kinesthetic teaching, teleoperation, and passive observation, while detailing key learning models such as gaussian mixture modeling, dynamic movement primitives, and reinforcement learning. By bridging gaps in the existing literature, the study evaluates 171 publications to provide applicability analyzes and optimization strategies to address real-world challenges such as demonstration noise and dynamic obstacles. [16] In contrast, Moreno et al. provide a systematic literature review according to the PRISMA (Preferred Reporting Items for Systematic Literature review and Meta-Analysis) 2020 [21] guidelines. By analyzing 61 research papers the study determines that while LbD has achieved a high level of maturity and experimental success in academic environments, it has not yet achieved widespread adoption in the assembly industry. The research identifies kinesthetic teaching and passive observation as the most prevalent methods for teaching assembly tasks, though it notes that most current solutions focus on trajectory reproduction and struggle with task complexity. Despite the promise of LbD to make robot programming accessible to non-experts and support the manufacturing shift toward HMLV, critical obstacles remain regarding generalization to new environments and the lack of standardized performance metrics. [8]

Table 1: Previous literature reviews in the area of learning-by-demonstration

| Title [source] | Review method | Content of research | Outlook | Scope AS | PO | SD |
|---|---|---|---|---|---|---|
| A Survey of Imitation Learning for Contact-Rich Tasks in Robotics [13] | Systematic organization of current research and identification of challenges in imitation learning for contact-rich tasks in robotics | 1. Collecting demonstrations<br>2. Learning methods<br>3. Applications | a) Design theory for dual-process hierarchical architectures<br>b) Multimodal sensing<br>c) Bridging the gap between the real world and simulation | ◕ | ◔ | ◔ |
| A Survey of Imitation Learning: Algorithms. Recent Developments, and Challenges [14] | Comprehensive review of imitation learning, embedded into a historical and logical framework | 1. Behavioral cloning<br>2. Inverse RL<br>3. Adversarial imitation learning<br>4. Imitation from observation<br>5. Challenges & imitation | a) Collect diverse and large scale demonstrations<br>b) Crossdomain learning<br>c) Cross-embodiement learning<br>d) Learn from imperfect demonstrations | ◑ | ◔ | ○ |
| Deep Generative Models in Robotics: A Survey on Learning from Multimodal Demonstrations [15] | Unified and comprehensive survey on deep generative models in robotics | 1. Problem formulation<br>2. Densitiy estimation models<br>3. Integrating generative models into robotics<br>4. Off-distribution generalization<br>5. Research challenges & outlook | a) Robot policies for long horizon tasks<br>b) Learning from video demonstrations<br>c) Learning from synthetic data<br>d) Learning from online interaction<br>e) Generalization | ◔ | ◕ | ○ |
| Learning from demonstration for autonomous generation of robotic trajectory: Status quo and forward-looking overview [16] | Two-step systematic literature review (identification, screening): definition of search-string, databases, journals/conferences, research field | 1. Demonstration modes<br>2. Learning models<br>3. Optimization methods for demonstrations and generalization<br>4. Approach validation and application<br>5. Discussion on other imitation learning models | a) Biological signals to support LbD<br>b) Optimize demonstration to eliminate human factors<br>c) New AI algorithms to reuse/enhance demonstrations in new industrial settings<br>d) Digital twin to integrate HRC<br>e) Establish standards for symbolic learning | ◕ | ◔ | ○ |
| Obstacles and opportunities for learning from demonstration in practical industrial assembly: a systematic literature review [8] | Systematic literature review, based on the PRISMA 2020 [21] guidelines | 1. Approaches in LbD for assembly<br>2. Main research areas<br>3. Training techniques<br>4. Learning behaviors<br>5. Obstacles hindering practical implementation | a) Proven practicability<br>b) Task complexity and diversity<br>c) Generalisation<br>d) Performance evaluation<br>e) Integration concepts | ● | ◑ | ◔ |

Ultimately, the scope fitness assessment shows that the combination of passive observation with a single demonstration for assembly tasks remains unaddressed in current reviews. This justifies the need for a new study to expore this unique intersection of LbD methods. The objective of this review is to answer a series of research questions regarding LbD using passive observation with a single demonstration, tackling assembly tasks:

1) *What assembly operations are learned and carried out by robots?*
2) *What hardware and which algorithms are used to perceive the environment?*
3) *How does the system control the robot based on the perceived information?*
4) *What are obstacles that hinder the deployment of LbD solutions in the assembly industry?*

## 3. Research methodology

The research methodology applied in this systematic literature review is a mixture of Moher et al. and Tranfield et al. [17, 18], successfully applied by other authors to draw insights from the scientific literature (e.g.[19, 20]). This review adheres to the PRISMA guidelines [21] and follows a four-stage process, as illustrated in Figure 1.

To identify publications, comprehensive queries are executed in the ScienceDirect (www.sciencedirect.com) and Scopus (www.scopus.com) databases. As illustrated in Figure 2, the search keywords are grouped into three distinct categories: *Learning-by-Demonstration* and synonyms as method, *robot* and similar words as reproduction system, and the broader field of *assembly* as application domain. Because there are a variety of synonyms and different definitions used in academia, similar words are grouped for the queries. In the eligibility phase, exclusion criteria are used to eliminate all non-related publications, see Figure 1 and Table 2. The term LbD and synonyms are based on the processual definition from Zhu and Hu (2018), defining three phases (observation, encoding, execution) within the LbD with the aim to make a robot learn an action based on a demonstration [22]. The term robot and similar words are defined based on the definition from the Verein Deutscher Ingenieure (Association of German Engineers), defining robots as "…universally applicable moving machines with several axes which are freely programmable (i.e. without mechanical intervention) and sensor driven regarding their movement sequence, course or angles. They can be equipped with grippers, tools or other means of production and are capable of accomplishing handling and/or production tasks." [23]. The term assembly is defined according to Lotter (2013) [24]. However, the reviewed literature considers an assembly operation on a subassembly sufficient.

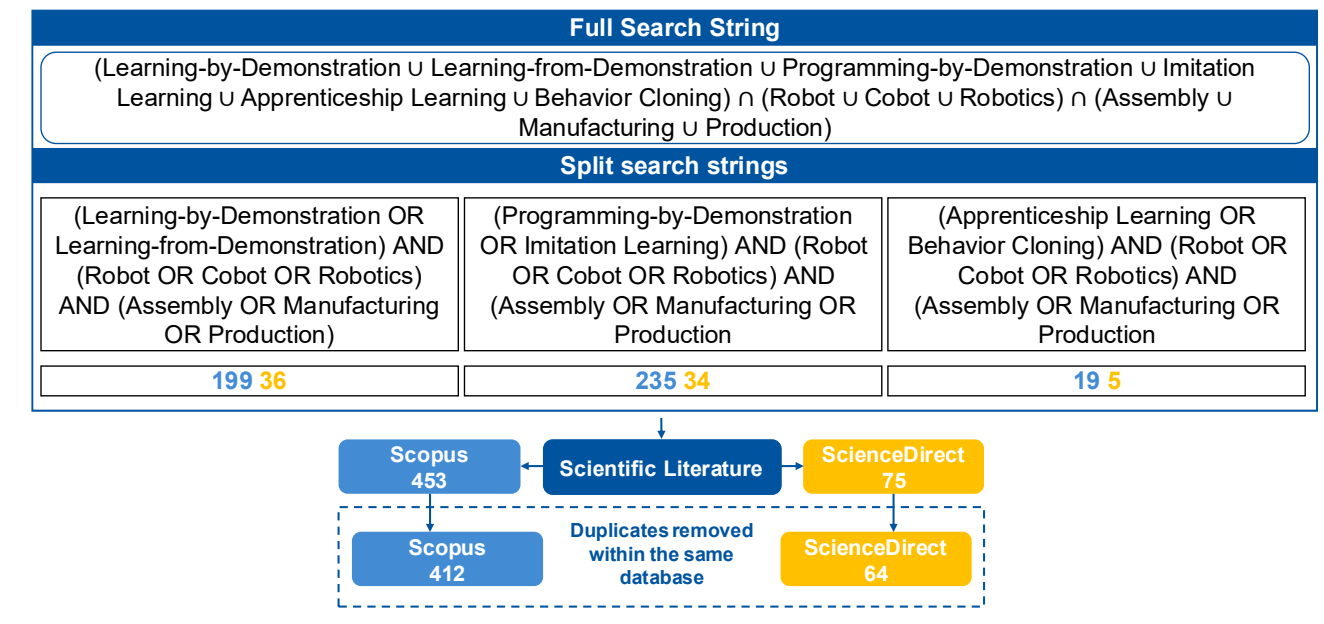

Figure 2: Search strings applied in this literature review

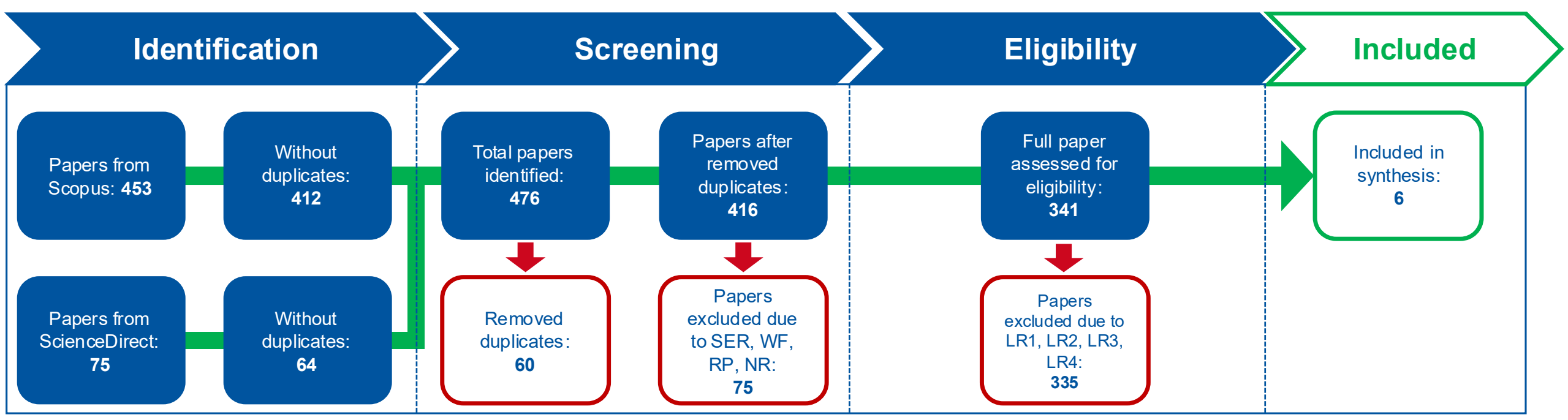

Figure 1: Four stage systematic literature review

Table 2: Inclusion and exclusion criteria for the four stage systematic literature review

| Inclusion / Excl. | Criteria | Criteria explanation |
|---|---|---|
| Inclusion | Closely related (CR) | • The research efforts of a paper are explicitly and specifically dedicated to learning-by-demonstration with passive observation for an industrial robot with a single demonstration, tackling (an) assembly task(s)<br>• Time Span: All papers available on 1st September 2025<br>• Subject area: Production, Manufacturing, Assembly<br>• Document type: Coference paper or article<br>• Source Type: Conference Proceedings or Journals or Book Chapter<br>• Language: English |
| Exclusion | Search engine reasons (SER) | A paper has only its title, abstract, and keywords in English, but not its full text |
| | Without full text (WF) | A paper without fulltext to assess |
| | Review papers (RP) | Review papers already discussed in Section 2 |
| | Non-related (NR) | A paper is not an academic article. For example, editorial materials, conference reviews, contents, or forewords |
| | Loosely related (LR) | A paper does not focus on the discussion, or problem solving of one-shot learning-by-demonstration with passive observation for industrial assembly. In which:<br>• LR1: The article does not propose a (novel) method for LbD to learn assembly operations<br>• LR2: The method requires something else than passive observing a human teacher<br>• LR3: The method requires something else than a single demonstration without a dedicated input device<br>• LR4: The method is not evaluated on an industrial robot executing the learned assembly operation(s) in the physical world |

In the screening phase, search results are consolidated into a spreadsheet and deduplicated. Title and abstract are screened to filter out irrelevant studies (SER = 6, WF = 32, RP = 19 and NR = 18). Subsequently in the eligibility phase, the full text of the remaining papers is assessed for eligibility based on criteria LR1 – LR4, ensuring that all the publications analyzed in the included phase (number of publications that do not meet the statement):

- Introduce a novel method for LbD to learn (an) assembly operation(s) (LR1 = 29)
- Use passive observation with human performed demonstration for data collection (LR2 = 98)
- Require a single demonstration without a dedicated input device (LR3 = 75)
- Evaluate the method with an industrial robot executing the learned assembly task(s) in the physical world (LR4 = 133)

## 4. Results

In the subsequent analysis, the identified papers are organized in relation to the research questions established in Section 2.

### *4.1 What assembly operations are learned and carried out by robots?*

The identified literature focuses extensively *on pick-and-place* tasks [25–30], which belong to *moving* and are categorized as *handling* assembly operation according to the taxonomy provided in [23, 24]. Within this corpus, certain studies investigate broader assembly functionalities. For instance, [26] incorporates *insertion* and *screwing*, representing both *put together* and therefor *joining* operations [24, 31]. Additionally, [30] details *pushing* operations, which constitute a specific subset of the *moving* assembly category [24].

### *4.2 What hardware and which algorithms are used to perceive the environment?*

All analyzed studies [25–30] rely on RGB-D imaging to record demonstrations. Hardware preferences are divided between the Microsoft Kinect [25, 30] and Intel RealSense [26, 27] platforms, with [28] opting for the ASUS Xtion Pro Live. The specific sensor implementation remains unspecified in [29]. Except for [25], which employs an eye-level configuration, all reviewed studies utilize a bird's eye perspective for camera mounting [26–30].

Perception strategies vary from deep learning frameworks to classical geometric analysis: [25] combines YOLO with Scale-Invariant Feature Transform [32] for object tracking and pose estimation, pretrained on the Pascal VOC 2007 and 2012, as well as finetuned on a custom dataset, whereas [26] and [30] leverage MediaPipe [33] for hand landmark extraction, with [26] incorporating an LSTM network for grasp categorization and [30] employing frequency-based filters and Hough lines to determine 6D poses. In contrast, [27] utilizes interactive HSV-based segmentation and Iterative Closest Point (ICP) registration - incorporating a 5x5 depth kernel to mitigate sensor noise - while [28] processes point clouds via a 5 mm voxel grid and RANSAC background removal, employing a weighted feature vector comprising color, size, and concave hull contours. Lastly, [29] facilitates tracking through a color descriptor and Softmax classifier pipeline.

### *4.3 How does the system control the robot based on the perceived information?*

The control strategies employed by the analyzed systems represent a spectrum of methodologies ranging from direct kinematic mapping to complex semantic reasoning and virtual simulations. [25] dynamically tracks the workpiece positions, which are transformed into the manipulator's coordinate frame and processed through inverse kinematic models to derive joint-specific control signals. In a more cognitively oriented approach, [26] employs a Finite State Machine (FSM) triggered by changes in grasp type and velocity, leveraging an OWL-based Knowledge Base and SPARQL queries to semantically translate human actions into robot-specific tasks via a skill model. [27] utilizes a mix of scaling and rule-based extraction, employing trajectory-based control by extracting via points from demonstrated object movements, represented by large turning angles calculated using the adjacent way-points to maintain trajectory's shape. Upon completion of the placement process, the system registers and saves the specific container, used as the designated destination for that object category in the Object Library. In contrast, [28] adopts a behavior policy governed by hierarchical FSMs that manage sequential object manipulation through eight predefined states, allowing for the dynamic allocation of object positions at runtime. Structurally, [29] transforms perceived data into an assembly graph to solve systems of equations for precise pose calculation, optimizing the assembly sequence from bottom-to-top while identifying "optimal fixing constraints" to maximize the manipulator's motion capability. Finally, [30] implements a Digital Twin (DT) pipeline that maps human movements to predefined robot primitives; this system utilizes a virtual environment to perform feasibility and singularity checks via bi-directional TCP communication before the validated control knowledge is transferred to the physical hardware.

*4.4 What are obstacles that hinder the implementation of LbD solutions in the assembly industry?*

Despite the potential of LbD through passive observation, several critical obstacles hinder its transition from laboratory settings to industrial assembly floors.

**Task variety.** A primary limitation is the narrow scope of investigated tasks; while the industry requires intricate, contact-rich operations, research remains largely fixated on basic pick-and-place movements, with only [26] exploring complex processes like screwing or insertion.

**High-level control architecture.** The narrow focus on pick-and-place tasks is further exacerbated by the absence of overarching, modular high-level control architectures, such as Behavior Trees [34], which are indispensable for managing task sequencing, reusability, and error recovery in dynamic industrial settings. While certain research, notably [26, 28], utilize FSMs to provide a logical framework, compared to Behavior Trees, FSMs are not fault-tolerant by design, often imposing a restrictive trade-off between system reactivity and architectural readability [35]. Consequently, Behavior Trees emerge as a superior choice for industrial control, offering a hierarchical and modular approach that improves reusability, scalability and readability as task complexity grows, yet they remain significantly underutilized in current LbD frameworks.

**Intuitive human-machine interface.** The deployment is further restricted by the absence of intuitive, usable interfaces designed for non-expert shop-floor operators, creating a high barrier to entry that necessitates constant developer intervention. Furthermore, evaluations are typically performed under idealized conditions using simplified components rather than real-world products, such as control cabinet assemblies [26], limiting their relevance to industrial settings.

**Perception.** The perception layer remains a significant technical bottleneck; users are often forced to choose between requiring specialized expertise in deep learning for object detection [25] or resorting to the manual, brittle tuning of parameters [27] which is sensitive to shifting lighting conditions. [26] shifts the focus from detecting objects to having objects in a calibrated trail, detecting only hands and grasping types instead of objects. However, this requires a dedicated and well-prepared working area for the robot, with the components being located at predefined locations, requiring plenty of manual work in changing the system to assemble new products variants.

Collectively, these factors maintain a substantial reality gap that prevents LbD from becoming a production-ready solution for modern automation in the factory of the future.

## 5. Conclusion

LbD has emerged as a solution for HMLV manufacturing, offering a way to drastically reduce the cost and complexity of robot (re-)programming. This systematic literature review specifically targeted the intersection of passive and one-shot learning to carry out assembly operations with industrial robots. Out of 478 identified papers, only six met the criteria for learning assembly operations based on passive observation with a single demonstration, highlighting a significant research gap in current literature. Existing studies rely heavily on RGB-D imaging (such as Microsoft Kinect or Intel RealSense) and bird's eye perspectives to record demonstrations. Perception strategies range from classical geometric analysis to deep learning frameworks like YOLO and MediaPipe, whereas control strategies vary from direct kinematic mapping to more complex semantic reasoning and Digital Twin pipelines. However most solutions remain limited to basic handling operations, such as pick-and-place tasks.

To transition LbD from laboratory environments to production-ready industrial floors, several critical challenges must be overcome. Future research needs to move beyond simplified benchmark components and basic pick-and-place movements to address the physical and visual complexities of real-world industrial products and contact-rich operations like screwing or insertion. A significant shift is required in high-level control architectures; replacing rigid Finite State Machines (FSMs) with modular and fault-tolerant Behavior Trees could improve system reactivity and scalability as task complexity grows. Furthermore, the industry requires more intuitive, usable human-machine interfaces designed for non-expert shop-floor operators to remove the barrier of constant developer intervention, especially in adapting the system to (perceive and assemble) new product variants in HMLV assembly environments.

## Acknowledgements

Funded by the Deutsche Forschungsgemeinschaft (DFG, German Research Foundation) under Germany's Excellence Strategy – EXC-2023 Internet of Production – 390621612. We thank Josefine Monnet for proofreading and helpful comments.